\documentclass[runningheads]{llncs}
\usepackage{graphicx}
\usepackage{comment}
\usepackage{amsmath,amssymb} 
\usepackage{color}
\usepackage{cases}
\usepackage{algorithm}
\usepackage{algorithmic}
\usepackage{overpic}

\usepackage{pgfplots}
\usepackage{pgf,tikz}
\usetikzlibrary{calc,patterns,angles,quotes,positioning}

\usepackage[width=122mm,left=12mm,paperwidth=146mm,height=193mm,top=12mm,paperheight=217mm]{geometry}

\newcommand{\mX}{\mathcal{X}}
\newcommand{\tq}{\tilde{q}}
\newcommand{\mW}{\mathcal{W}}
\newcommand{\bbR}{\mathbb{R}}

\begin{document}
\pagestyle{headings}
\mainmatter
\def\ECCVSubNumber{1676}  

\title{Hamiltonian Dynamics for Real-World \\Shape Interpolation} 

%
\author{Marvin Eisenberger, Daniel Cremers}
\authorrunning{F. Author et al.}
%
\institute{
Technical University of Munich, Garching, Germany}
\maketitle

\begin{abstract}
We revisit the classical problem of 3D shape interpolation and propose a novel, physically plausible approach based on Hamiltonian dynamics. While most prior work focuses on synthetic input shapes, our formulation is designed to be applicable to real-world scans with imperfect input correspondences and various types of noise. To that end, we use recent progress on dynamic thin shell simulation and divergence-free shape deformation and combine them to address the inverse problem of finding a plausible intermediate sequence for two input shapes. In comparison to prior work that mainly focuses on small distortion of consecutive frames, we explicitly model volume preservation and momentum conservation, as well as an anisotropic local distortion model. We argue that, in order to get a robust interpolation for imperfect inputs, we need to model the input noise explicitly which results in an alignment based formulation. Finally, we show a qualitative and quantitative improvement over prior work on a broad range of synthetic and scanned data. Besides being more robust to noisy inputs, our method yields exactly volume preserving intermediate shapes, avoids self-intersections and is scalable to high resolution scans. 
\end{abstract}

\begin{figure*}[!b]
    \centering
    \includegraphics[width=\linewidth]{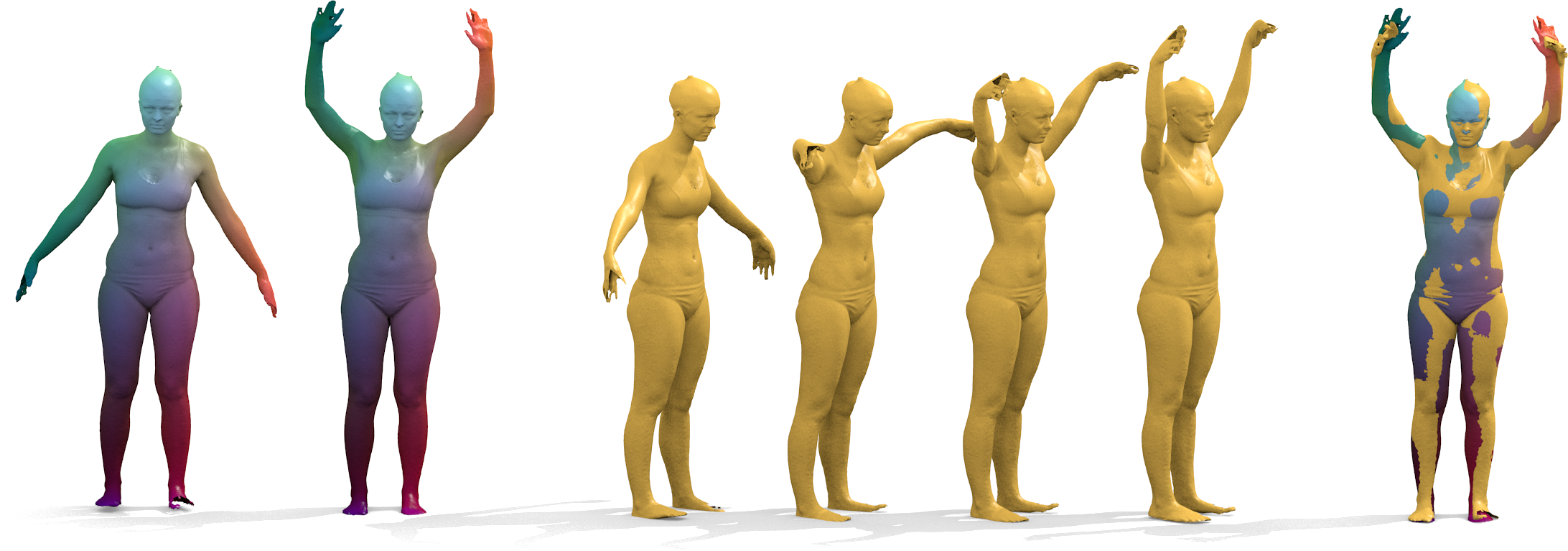}
    \caption{An example interpolation (middle) on real scans from the FAUST dataset \cite{Bogo:CVPR:2014} and the final overlap (right). Here, the input correspondences were computed with Deep Functional Maps \cite{litany2017deep} (left).}
    \label{fig:faust_teaser}
\end{figure*}

\section{Introduction}

Modeling realistic deformations of 3D shapes is at the heart of many computer vision applications. The central motivation in this context is to give meaning to sparse observations of a dynamically moving 3D object.
Depending on the application, these measurements are given in the form of a point cloud, a triangle mesh, a voxel grid or a signed distance function. 

In many cases, the sampling is not consistent over time and finding commonalities between observations is not a trivial task. While there are a lot of approaches that try to fuse scanned data for 3D reconstruction \cite{newcombe2015dynamicfusion,slavcheva2017killingfusion,slavcheva2018sobolevfusion}, relatively few work was dedicated to modeling the temporal transformation of the observed object directly. 
In this work, we revisit this classical challenge of 3D shape interpolation. Although there exists a multitude of elegant formulations, we will show that a lot of these approaches are mainly designed for synthetic shapes and therefore lack robustness to noisy real-world measurements. 

The classical formulation is to define an interpolation as a sequence of shapes with minimal local distortion between consecutive frames \cite{brandt2016geometric,heeren2012time,kilian2007geometric}. While this is undoubtedly a reasonable assumption, it does not suffice in practice to account for the peculiarities of real-world data. For synthetic 3D objects, the ground-truth correspondences are typically known. For real scans, on the other hand, we need to first estimate them, e.g. by using a shape matching method. In practice, the resulting correspondences are not perfect and contain both outliers and fine-scale noise. This is problematic for an interpolation method that minimizes the local distortion between neighboring frames, because the noise from the faulty correspondences tends to distort the local geometry throughout the whole sequence. Moreover, most classical approaches do not model the global geometry of an object which can lead to artifacts like self-intersections. 

\subsubsection{Contribution}
We propose a novel framework for real-world shape interpolation that is systematically derived from Hamiltonian dynamics.  It resolves the above challenges by introducing additional, physically plausible modeling assumptions like volume preservation and momentum conservation. More specifically, we formulate shape interpolation as the inverse problems of a dynamic thin shell simulation. The Eulerian time-varying deformation fields are represented in a low rank manner which allows us to build volume preservation directly into our model. In qualitative and quantitative experiments, we demonstrate that our method gives rise to high-quality interpolations for real-world inputs.

\section{Related work}

Shape interpolation has a long tradition in computer graphics. Originally it was developed for planar shapes \cite{alexa2000asrigidas,michor2003riemannian,sederberg19932} with \cite{chen2013planar} being a more recent formalism.
A common approach for 3D surfaces is to define an interpolating trajectory as a geodesic in some higher dimensional shape space \cite{brandt2016geometric,heeren2014exploring,heeren2016shellsplines,wirth2011continuum,wirth2011shapespace}. Most of these methods use some kind of deformation measure and then optimize for a sequence such that the local distortion between any two consecutive shapes is low. In \cite{kilian2007geometric} this is done with an as-killing-as-possible energy and in \cite{heeren2012time} with a discrete shell energy motivated by \cite{grinspun2003discrete}. Other popular examples of non-linear shape deformation are PriMo \cite{botsch2006primo} and as-rigid-as-possible \cite{sorkine2007rigid}. For a more thorough introduction to shape spaces, we refer the reader to the book of Younes \cite{younes2010shapes}.

An alternative approach to shape interpolation is to interpolate intrinsic quantities like dihedral angles before reconstructing the extrinsic geometry \cite{alexa2000asrigidas,baek2015isometric,xu2005poisson}. One class of such intrinsic quantities are rotation-invariant or differential coordinates \cite{alexa2003differential,lipman2004differential,lipman2005linear,sassen2019geometric}.

Sometimes shape deformation is stated as the time-dependent gradient flow wrt. some surface functional. Typically these functionals promote a smooth flow \cite{charpiat2005approximations,charpiat2007generalized,eckstein2007generalized,sundaramoorthi2007sobolev} but most of these methods focus on shape matching with less emphasis on the quality of the intermediate shapes.

Recently, more and more work was dedicated to processing collections of shapes in order to make interpolation more efficient. This can e.g. be achieved by constructing a low-dimensional subspace of admissible poses \cite{allen2006learning,fletcher2004principal,heeren2018principal,sumner2005mesh,zhang2015shell}. In practice, this greatly helps to reduce the computational cost of shape interpolation and even allows for interactive applications \cite{von2016optimized}.

A common assumption of interactive shape deformation modeling is volume preservation. This can be obtained by defining a deformation as the flow of a divergence-free Eulerian vector field \cite{angelidis2006swirling,von2006vector}. Recently, \cite{eisenberger2019divergence} extended this idea by constructing a divergence-free vector field basis that can be used to interpolate 3D objects. We will make use of this vector field representation and additionally formulate shape interpolation as the inverse problem of a dynamical thin shell simulation. The forward simulation corresponding to this is a well-known problem in computer graphics \cite{muller2007position} with applications like cloth \cite{goldenthal2007efficient} or fluid \cite{macklin2013position} simulation. A recent formulation of this problem that is akin to our approach is projective dynamics \cite{bouaziz2014projective,brandt2018hyper}. Here, the Lagrangian gradient
flow of a dynamical system is restated using the variational form of implicit Euler integration from \cite{martin2011example} which leads to an efficient and extremely robust thin shell simulation.

\section{Background}

We briefly review important preliminary work on shape deformation and interpolation of non-rigidly deforming 3D objects. In this work, we focus on surface-based models like point clouds and 3D meshes. This allows for a compact representation and is in coherence with the output of real-world sensors. In particular, the set of observations $p=(p_1,\dots,p_n)^\top\in\bbR^{n\times 3}$ consists of $n$ points sampled from a two-dimensional Riemannian manifold $\mX$. Depending on the application, these points are either part of a triangle mesh or embedded in a (knn-)graph. 

\subsection{Physical assumptions for shape deformation}

In order to find similarities between two non-rigid poses of an object, it helps to model geometric assumptions about the expected deformations directly. We review two common assumptions, namely small local distortions and volume preservation.

\paragraph{Local distortion} A popular deformation energy to quantify the distortion between $p$ and a deformed counterpart $p^*\in\bbR^{n\times 3}$ is the as-rigid-as-possible (arap) energy \cite{sorkine2007rigid}:
\begin{equation}
    \label{eq:arapenergy}
    \mW_\mathrm{arap}\bigl(p,p^*;(R_i)_{1\leq i\leq n}\bigr)=\frac{1}{2}\sum_{i=1}^n\sum_{j\in\mathcal{N}(i)}\bigl\|R_i\bigl(p_j-p_i\bigr)-\bigl(p_j^*-p_i^*\bigr)\bigr\|_2^2.
\end{equation}
The assumption behind this functional is that the local deformation of the geometry in the neighborhood $\mathcal{N}(i)$ of every vertex $p_i$ is approximately rigid. I.e. one can find a rotation matrix $R_i\in SO(3)$ that approximately captures the transformation of the neighboring edges $p_j-p_i$. In turn, deviations of the deformation $p^*$ from the approximate rigidity are penalized. The neighborhood $\mathcal{N}(i)$ for a given vertex $i$ is defined as the set of adjacent vertices $j$ to $i$.

There are multiple popular alternatives with the same flavor as $\mW_{arap}$, including PriMo \cite{botsch2006primo}, discrete shells \cite{grinspun2003discrete} and as-killing-as-possible \cite{solomon2011killing}. Most techniques penalize deformations of the local geometry and each one of them has certain advantages. In our formulation, we choose the arap energy because it is applicable directly for point clouds and because the optimization for $p^*$ and $R_i$ can be done efficiently in closed form.

\paragraph{Volume preservation} Another common assumption for shape deformation is that the volume of the observed object is preserved over time \cite{von2006vector}. This can be obtained by prescribing that the deformation is the flow induced by an underlying Eulerian deformation field $v:\bbR^3\to\bbR^3$ which is divergence-free $\mathrm{div}(v)=0$. Recently, \cite{eisenberger2019divergence} proposed a formulation of a coarse-to-fine vector field basis that has the volume preservation built in as a hard constraint. A flow field $v$ is then obtained as the linear combination of a finite subset of those divergence-free basis functions:
\begin{equation}
    \label{eq:divfreebasis}
    \mathrm{v}(x;c)=\sum_{k=1}^{K}c_k\phi_k(x)\text{, where }\mathrm{div}(\phi_k)=0
\end{equation}
These deformation fields $v$ are exactly volume preserving because the divergence is a linear operator, see \cite{eisenberger2019divergence} for more details. In practice, a relatively small number $K\approx 1000$ of coefficients $c=(c_1,\dots,c_K)^\top\in\bbR^K$ suffices to represent arbitrary smooth, volume preserving vector fields $v$. 
We make use of this compact representation in this work. However, while \cite{eisenberger2019divergence} only considered stationary vector fields $v(x)$, in this work we consider time-dependent vector fields $v(t,x)$ in order to account for more complex shape variation.

\subsection{Shape interpolation}

Computing an interpolation of two 3D objects $p=p^{(0)}$ and $q=p^{(T)}$ is a common problem in computer graphics and Vision. In general, it is not a well-defined problem because there are typically infinitely many conceivable paths between $p^{(0)}$ and $p^{(T)}$. Therefore, we need to make additional assumptions about plausible sequences like small local distortions or volume preservation. The common way to do this is to define a deformation energy for the whole, time-discrete sequence $p^{(0)},\dots,p^{(T)}$ of intermediate shapes \cite{brandt2016geometric,heeren2016shellsplines,heeren2012time,huber2017smooth,kilian2007geometric}:
\begin{equation}
    \label{eq:nsequenceenergy}
    E\bigl(p^{(1)},\dots,p^{(T-1)}\bigr)=\sum_{t=0}^{T-1}\mW\bigl(p^{(t)},p^{(t+1)}\bigr).
\end{equation}
Here, $\mW$ is some local distortion measure like $\mW_\mathrm{arap}$ from Eq.~\eqref{eq:arapenergy}. For symmetry reasons, the optimization is commonly done jointly for both the standard and the inverse sequence $p^{(T)},\dots,p^{(0)}$. W.l.o.g. we will consider the time interval $[0,t_\mathrm{max}]=[0,1]$ which leads to a discrete step size $\tau=\frac{1}{T}$.
\begin{figure*}[!b]
    \centering
    \begin{overpic}
        [width=\linewidth]{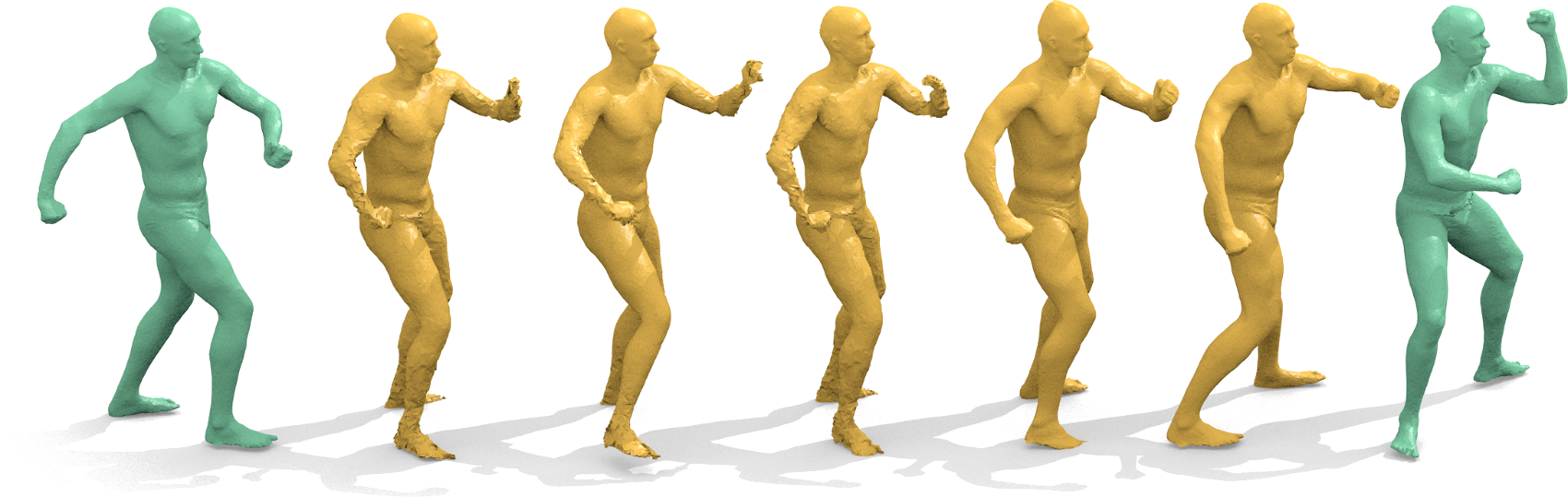}
        \put(24,33){\cite{kilian2007geometric}}
        \put(38.5,33){\cite{sorkine2007rigid}}
        \put(52,33){\cite{heeren2012time}}
        \put(66,33){\cite{eisenberger2019divergence}}
        \put(78,33){Ours}
    \end{overpic}
    \caption{A qualitative comparison of our approach with other popular shape interpolation methods. Here, we display the intermediate shapes at $t=0.5$ for an example pair from SCAPE~\cite{anguelov2005scape} with correspondences from BCICP~\cite{ren2018orientation}. Like us, \cite{eisenberger2019divergence} solves for an approximate alignment formulated as an IVP but the stationary vector field leads to slight distortions of the geometry (e.g. at the head and right arm). The other methods \cite{heeren2012time,kilian2007geometric,sorkine2007rigid} solve a BVP and in certain areas the high frequency noise of the correspondences from BCICP leads to a severely degenerate geometry.}
    \label{fig:scape_compare}
\end{figure*}
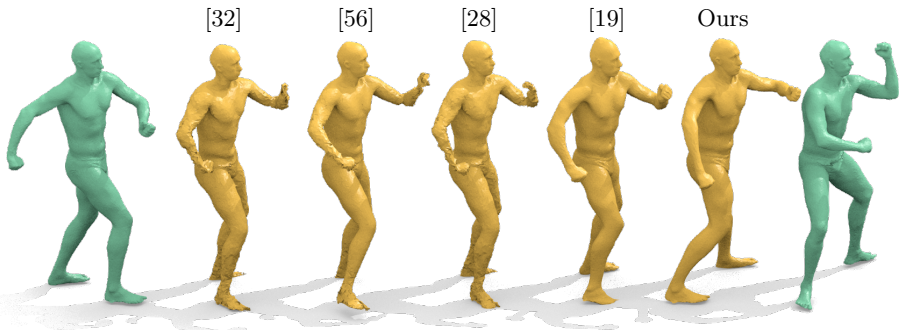
\section{Interpolation of real-world objects}\label{sec:shapeinterpolationrealworld}
The implicit assumption behind most shape interpolation approaches is that the exact point-to-point correspondences between the two input surfaces $p$ and $q$ are known. While there is a lot of synthetic data where this is feasible, for scanned data the sampling of two given objects is typically not consistent, even if they approximate the same real-world surface $\mX$. Not even the number of points of the two surfaces $p\in\bbR^{n\times 3}$ and $q\in\bbR^{m\times 3}$ is necessarily the same in the most general case. In order to compute an interpolation for this type of input data, we need to first estimate the surface correspondences between $p$ and $q$.

Computing shape correspondences is a problem in itself and there is a variety of methods that focus on shape matching, either in the classical sense \cite{eisenberger2019smooth,kim11,melzi2019zoomout,ovsjanikov2012functional,ren2018orientation,kernel17} or using machine learning \cite{boscaini2016learning,groueix20183dcoded,litany2017deep,masci2015geodesic,monti2017geometric,rodola-cvpr14}. The output of those methods is a point-to-point assignment of the surface $p\in\bbR^{n\times 3}$ to $q\in\bbR^{m\times 3}$ which can be represented with a matrix $\mathrm{\Pi}\in\{0,1\}^{n\times m}$. 
In principle, we can now transfer the points and neighborhood information from $p$ to $q$ and apply a classical interpolation method like \cite{kilian2007geometric} or \cite{heeren2012time} to $p\in\bbR^{n\times 3}$ and $\mathrm{\Pi} q\in\bbR^{n\times 3}$. However, in practice the correspondences $\mathrm{\Pi}$ are not perfect and contain faulty or noisy matches. We found that most interpolation methods that assume perfect correspondences are not very robust to fine-scale noise, see Figure~\ref{fig:scape_compare}. 

One possible way to make interpolation feasible for real scans is to acknowledge that the given matching $\mathrm{\Pi}$ is not perfect and to build this stochastic discrepancy directly into our model. In particular, we add Gaussian random noise $\eta$ to the vertex position of the second shape $\mathrm{\Pi} q$:

\begin{equation}
    \label{eq:gaussnoise}
    \tq:=\mathrm{\Pi} q+\eta\text{, with }\eta\sim\mathcal{N}(0,\sigma).
\end{equation}

Instead of finding intermediate shapes by solving a boundary value problem (BVP) as outlined in Eq.~\eqref{eq:nsequenceenergy}, we can then define an initial value problem (IVP) similar to \cite{eisenberger2019divergence}. In particular, we will optimize for a sequence $p^{(0)},\dots,p^{(T)}$ with $p=p^{(0)}$ and $p^{(T)}=\tq\approx \mathrm{\Pi} q$.

\section{From Hamiltonian dynamics to Eulerian-Lagrangian shape interpolation}

\begin{figure*}[!b]
    \centering
    \begin{overpic}
        [width=\linewidth]{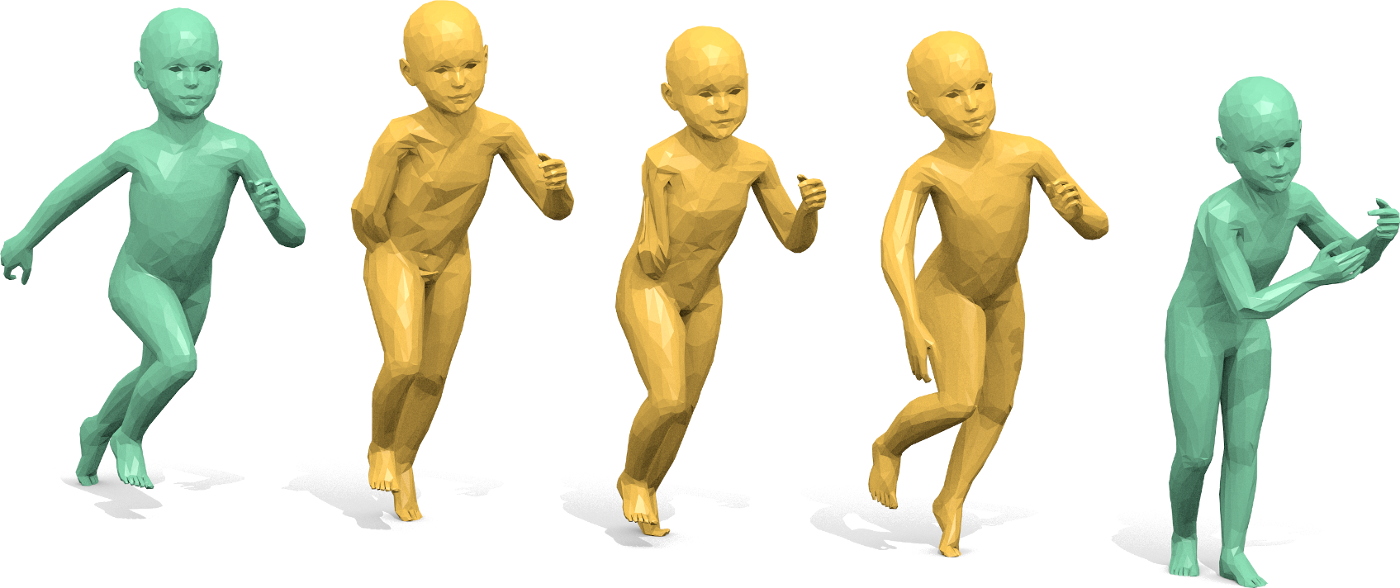}
        \put(29,44){\cite{heeren2012time}}
        \put(48,43){\cite{kilian2007geometric}}
        \put(65,42){Ours}
    \end{overpic}
    \caption{A pair of synthetic shapes with ground-truth correspondences from the KIDS dataset \cite{rodola-cvpr14} for which we show the intermediate shapes at $t=0.5$. This example shows that many classical methods like \cite{kilian2007geometric} or \cite{heeren2012time} cannot detect self-intersections of different subparts. Here, the optimal path that minimizes a local distortion metric makes the right arm of the kid move through itself. Our method, on the other hand, avoids self-intersections by design: All deformations are expressed as a divergence-free Eulerian field, therefore the resulting flow has to be globally consistent in the sense that two close parts cannot have contradictory motions.}
    \label{fig:self_intersect}
\end{figure*}

In this work, we model the motions of objects in an inertial frame of reference as a physical phenomenon that is governed by three aspects: internal forces, momentum conservation and volume preservation. 
Most existing interpolation techniques model internal forces in some way, yet they omit the momentum conservation and volume preservation. Without momentum conservation, the intermediate objects can be plausible but in many cases the motions lack temporal coherence. The volume preservation helps to constrain the optimization and prevents self-intersections, see Figure~\ref{fig:self_intersect}. 
Our formulation combines the strengths of volume preserving fields \cite{eisenberger2019divergence} and projective dynamics \cite{bouaziz2014projective} with those of classical interpolation methods \cite{heeren2012time,kilian2007geometric}. 

\subsection{Deformation model}

We systematically derive the evolution of a surface as a physical system from the Hamiltonian energy given by:
\begin{equation}
    \mathcal{H}(p,v)=\frac{1}{2}\|v\|^2_2+\mathcal{W}(p).
\end{equation}
This energy consists of a kinetic energy term that models momentum conservation (with unit mass per point) and some potential energy component $\mathcal{W}$ that penalizes intrinsic distortions. The principles of Hamiltonian mechanics now prescribe how this system evolves over time:
\begin{equation}
\label{eq:hamiltoniansystem}
\begin{cases}
    \dot{p}=\hspace{8pt}\frac{\mathrm{d}\mathcal{H}}{\mathrm{d}v}=v.\\[1mm]
    \dot{v}=-\frac{\mathrm{d}\mathcal{H}}{\mathrm{d}p}=-\nabla\mathcal{W}(p).
\end{cases}
\end{equation}
We couple this with the volume preservation assumption by constraining $v$ to the low rank vector field representation from Eq.~\eqref{eq:divfreebasis}. This allows us to model displacements of a shape $p(t)=\bigl(p_1(t),\dots,p_n(t)\bigr)^\top$ at time $t$ with only $K\ll n$ degrees of freedom:
\begin{equation}
    \label{eq:pointdisplacement}
    \dot{p}_i(t)=\mathrm{v}\bigl(p_i(t);c(t)\bigr)=\sum_{k=1}^{K}c_k(t)\phi_k\bigl(p_i(t)\bigr).
\end{equation}
Besides providing a compact representation, this approach builds volume preservation directly into the deformation model, because $\mathrm{div}(v)=0$. In \cite{eisenberger2019divergence}, the authors model shape deformations in a similar way but with a stationary vector field $\mathrm{v}\bigl(x;c(t)\bigr)=\mathrm{v}\bigl(x;c\bigr)$. This leads to a well-constrained optimization problem with only $K$ degrees of freedom $c_1,\dots,c_K$ but it is also restrictive and lacks expressivity. Instead of using a constant vector field, following Eq.~\eqref{eq:hamiltoniansystem}, we define a dynamic flow $v(t,x)=\mathrm{v}(x;c(t))$:
\begin{equation}
    \label{eq:velocityupdate}
    \begin{cases}
        \dot{v}\bigl(t,p(t)\bigr)=-\nabla\mathcal{W}\bigl(p(t)\bigr).\\[1mm]
        \mathrm{div}(v)=0.
    \end{cases}
\end{equation}
In our formulation, the internal forces are defined as the negative gradient of our anisotropic as as-rigid-as-possible potential $\mathcal{W}$ which we define in the next chapter.

\subsection{Anisotropic as-rigid-as-possible deformation}

\begin{figure}
    \centering
    \includegraphics[width=\linewidth]{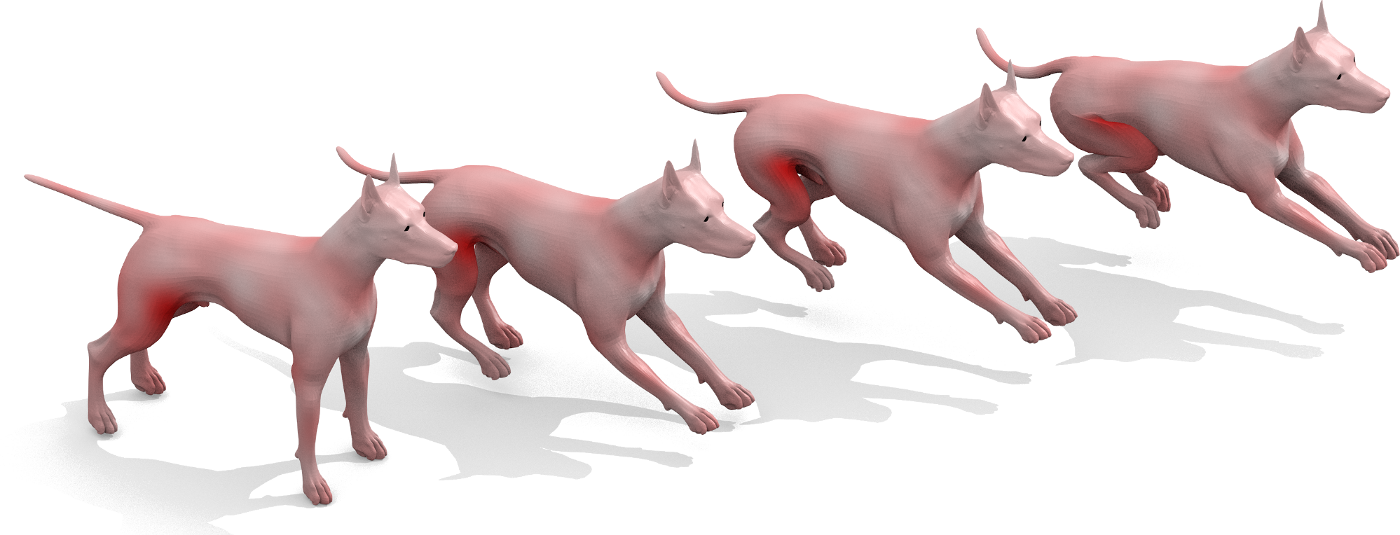}
    \caption{An example from TOSCA where we color code the log-determinant of the covariance matrix $\log\det(\Sigma_i)$ for each point $p_i$. Red stands for a low value which corresponds to a high local distortion. This shows how our anisotropic as-rigid-as-possible energy \eqref{eq:anisotropicarapenergy} automatically adapts to objects consisting of inhomogeneous parts. Certain regions like joints allow for more local distortion throughout the sequence than others. Notice the difference between the hind legs, the head and the rest of the body.}
    \label{fig:anisotropic_dog}
\end{figure}

For most 3D objects, not all parts are behaving similar in terms of local distortions. For example, regions near joints of a human body allow for more movement than most other parts of the surface. The classical as-rigid-as-possible potential that we reviewed in Eq.~\eqref{eq:arapenergy} penalizes distortions of the geometry uniformly in all directions and equal for all parts of the considered object. We generalize this idea and introduce an anisotropic as-rigid-as-possible energy:
\begin{equation}
    \label{eq:anisotropicarapenergy}
    \mW\bigl(p(t);(R_i)_{i},(\Sigma_i)_{i}\bigr)=\frac{1}{2}\sum_{i=1}^n\sum_{j\in\mathcal{N}(i)}\bigl\|\bigl(p_j(0)-p_i(0)\bigr)-R_i^\top\bigl(p_j(t)-p_i(t)\bigr)\bigr\|_{\Sigma_i}^2.
\end{equation}
In this context, $\|\cdot\|_{\Sigma_i}$ denotes the standard Mahalanobis norm \cite{mclachlan1999mahalanobis} with an unknown covariance matrix $\Sigma_i\in\bbR^{3\times 3}$. This energy $\mW$ allows our model to adapt the appropriate local behavior during the optimization, see Figure~\ref{fig:anisotropic_dog} for an example. Moreover, the distortion is always computed in the reference frame of the first pose $p(0)$. This means that we only need to compute the distortion model of $p(0)$ and therefore we only need one local distortion matrix per vertex $\Sigma_i$ for the whole sequence.

\subsection{Time discretization}

In the time-discrete setting, we can approximate Eq.~\eqref{eq:pointdisplacement} and Eq.~\eqref{eq:velocityupdate} using an implicit Euler intergration scheme:
\begin{subnumcases}{}
    p^{(t+1)}=p^{(t)}+\tau v^{(t+1)}.\\
    v^{(t+1)}=v^{(t)}-\tau \nabla\mathcal{W}\bigl(p^{(t+1)}\bigr).\\
    \mathrm{div}\bigl(v^{(t+1)}\bigr)=0.
\end{subnumcases}
This is a Eulerian-Lagrangian scheme: The velocity field is represented on the surface $v^{(t)}\in\bbR^{n\times 3}$ but the divergence-free condition $\mathrm{div}\bigl(v^{(t+1)}\bigr)=0$ is Eulerian.
In order to make this interaction tractable, we will use the divergence-free vector field representation from Eq.~\eqref{eq:divfreebasis} and combine it with the variational form of implicit Euler integration introduced in \cite{martin2011example}. This allows us to restate this scheme as an optimization problem in terms of the vector field coefficients $c\in\bbR^K$:
\begin{subnumcases}{\label{eq:scheme}}
    c^{(t+1)}=\underset{c,R}{\arg\min}\biggl\|\mathrm{v}\bigl(p^{(t)};c\bigr)-\bar{v}^{(t)}\biggr\|_F^2+\mW\biggl(p^{(t)}+\tau \mathrm{v}\bigl(p^{(t)};c\bigr);R,\Sigma\biggr).\label{eqsub:cupdate}\\
    v^{(t+1)}_i=\mathrm{v}\bigl(p_i^{(t)};c^{(t+1)}\bigr)=\sum_{k=1}^{K}c_k^{(t+1)}\phi_k\bigl(p_i^{(t)}\bigr).\\
    p^{(t+1)}=p^{(t)}+\tau v^{(t+1)}.\\
    \bar{v}^{(t+1)}=2v^{(t+1)}-v^{(t)}.\label{eqsub:extrapolate}
\end{subnumcases}
We refer the interested reader to \cite{martin2011example} and \cite{bouaziz2014projective} for more details on how this scheme is derived. The update of the coefficients $c$ in \eqref{eqsub:cupdate} can be computed using Gauss-Newton optimization. We use an additional extrapolation step \eqref{eqsub:extrapolate} to get a better prediction of the velocity $v^{(t+2)}$ which we justify in the following:
\begin{theorem}\label{thm:errororder}
For continuously differentiable vector fields, the extrapolation step \eqref{eqsub:extrapolate} of Algorithm~\ref{eq:scheme} yields an estimate $\bar{v}^{(t+1)}$ of $v^{(t+2)}$ with an error of order $\mathcal{O}(\tau^2)$. For the alternative scheme without step \eqref{eqsub:extrapolate} it is $\mathcal{O}(\tau)$.
\end{theorem}
This result implies that \eqref{eqsub:extrapolate} leads to a qualitative improvement because a better estimate $\bar{v}^{(t+1)}\approx v^{(t+2)}$ provides a more faithful approximation in the next update step \eqref{eqsub:cupdate} of $c$. See Appendix~\ref{appendix:proof1} for a proof of Thm.~\ref{thm:errororder}.

\subsection{Interpolation algorithm}

We will now use the scheme \eqref{eq:scheme} from last chapter to define an interpolation algorithm for two given shapes $p$ and $q$. In each iteration, we initialize the scheme with $p^{(0)}:=p$ and the unknown variables $c^{(0)}:=\hat{c}$ and $(\hat{\Sigma}_i)_{1\leq i\leq n}$. We then compute the deformed shapes $p^{(0)},\dots,p^{(T)}$ according to our scheme \eqref{eq:scheme}. Overall, this forward pass can be summarized as the differentiable solution operator $\mathcal{S}$:
\begin{equation}
    \mathcal{S}:\begin{cases}
    \bbR^K\times\bbR^{n\times3\times3}\to\bbR^{n\times 3\times (T+1)}.\\
    \bigl(\hat{c},\hat{\Sigma}\bigr)\mapsto \bigl(p^{(0)},\dots,p^{(T)}\bigr).
    \end{cases}
\end{equation}
The goal is now to find the input parameters $\hat{c}$ and $\hat{\Sigma}$ that lead to a tight alignment of the deformed shape $p^{(T)}$ with $q$ in accordance with Eq.~\eqref{eq:gaussnoise}. Together with our regularizer $\mW$ from Eq.~\eqref{eq:anisotropicarapenergy} this leads to the following energy:
\begin{equation}
    E\bigl(p^{(0)},\dots,p^{(T)};\mathrm{\Pi}\bigr):=\frac{1}{2\sigma^2}\bigl\|p^{(T)}-\mathrm{\Pi} q\bigr\|^2+\sum_{t=0}^{T}\mW\bigl(p^{(t)}\bigr).
\end{equation}
Putting everything together, we can derive the following algorithm:
\begin{algorithm}[H]
\caption{Volume preserving shape interpolation.}
\begin{algorithmic}
\label{alg:interpolation}
\REQUIRE $p\in\bbR^{n\times 3}$, $q\in\bbR^{m\times 3}$
\STATE $\hat{c} \leftarrow 0\in\bbR^K$
\STATE $\hat{\Sigma}_i \leftarrow \mathrm{Id}_3\in\bbR^{3\times 3}$
\STATE $\mathrm{\Pi} \leftarrow \text{match\_shapes}(p,q)\in\{0,1\}^{n\times m}$
\FOR{$i=1,\dots,N_{\mathrm{it}}$}
\STATE $\bigl(\hat{c},\hat{\Sigma}\bigr)\leftarrow\bigl(\hat{c},\hat{\Sigma}\bigr)-\gamma\nabla E\bigl(\mathcal{S}(\hat{c},\hat{\Sigma});\mathrm{\Pi}\bigr)$
\ENDFOR
\RETURN{$\bigl(p^{(0)},\dots,p^{(T)}\bigr):=\mathcal{S}\bigl(\hat{c},\hat{\Sigma}\bigr)$}
\end{algorithmic}
\end{algorithm}

In our implementation, we use a modern automatic differentiation toolbox to compute the gradient $\nabla E\circ S$ wrt. ${(\hat{c},\hat{\Sigma})}$ in Algorithm~\ref{alg:interpolation}. The choice of algorithm to compute the input correspondences $\mathrm{\Pi}$ is not further specified here because it is more or less arbitrary. We show various different possibilities in our experiments.

\begin{figure*}[!b]
    \centering
    \includegraphics[width=\linewidth]{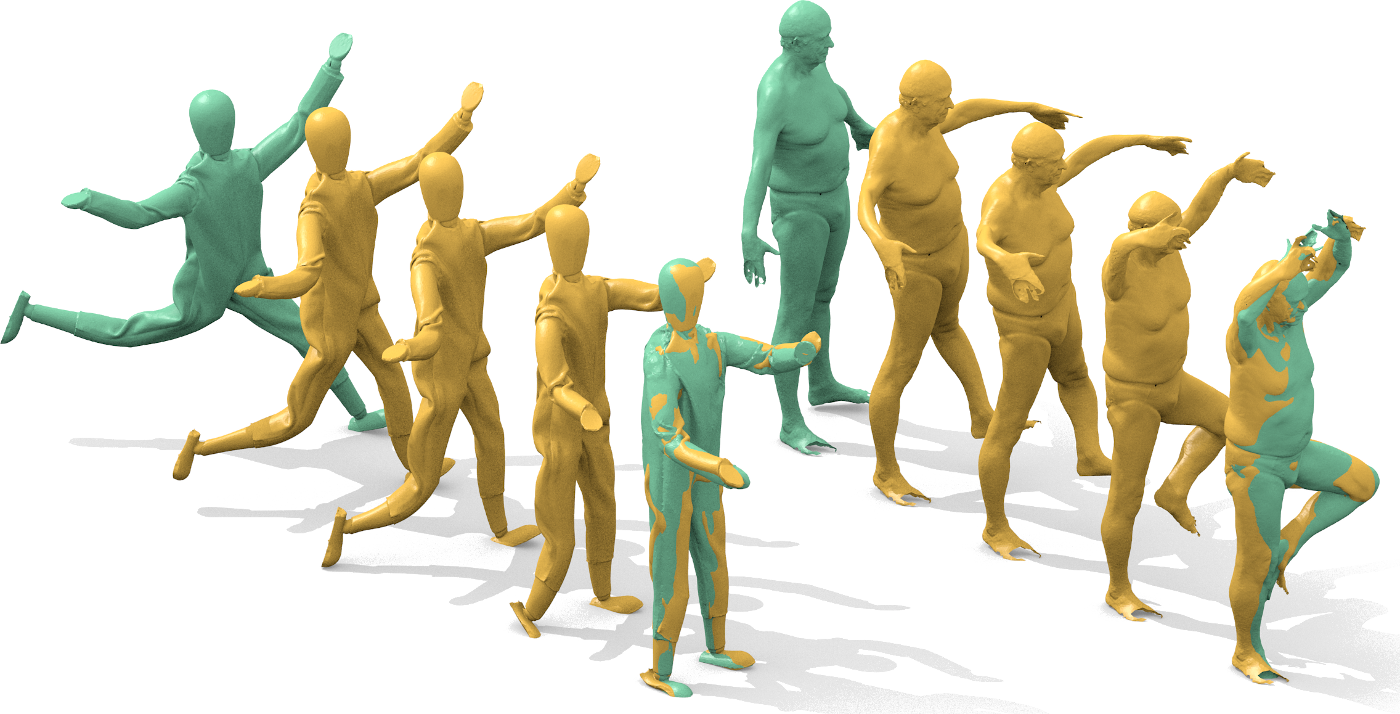}
    \caption{Two interpolated sequences for real scans of a puppet from SHREC'19 Isometry \cite{dyke2019shrec} and a human from FAUST \cite{Bogo:CVPR:2014} with input correspondences from Smooth Shells \cite{eisenberger2019smooth} and Deep Functional Maps \cite{litany2017deep} respectively.}
    \label{fig:scans}
\end{figure*}

\section{Experiments}

\begin{figure*}[!b]
    \centering
    \begin{overpic}
        [width=\linewidth]{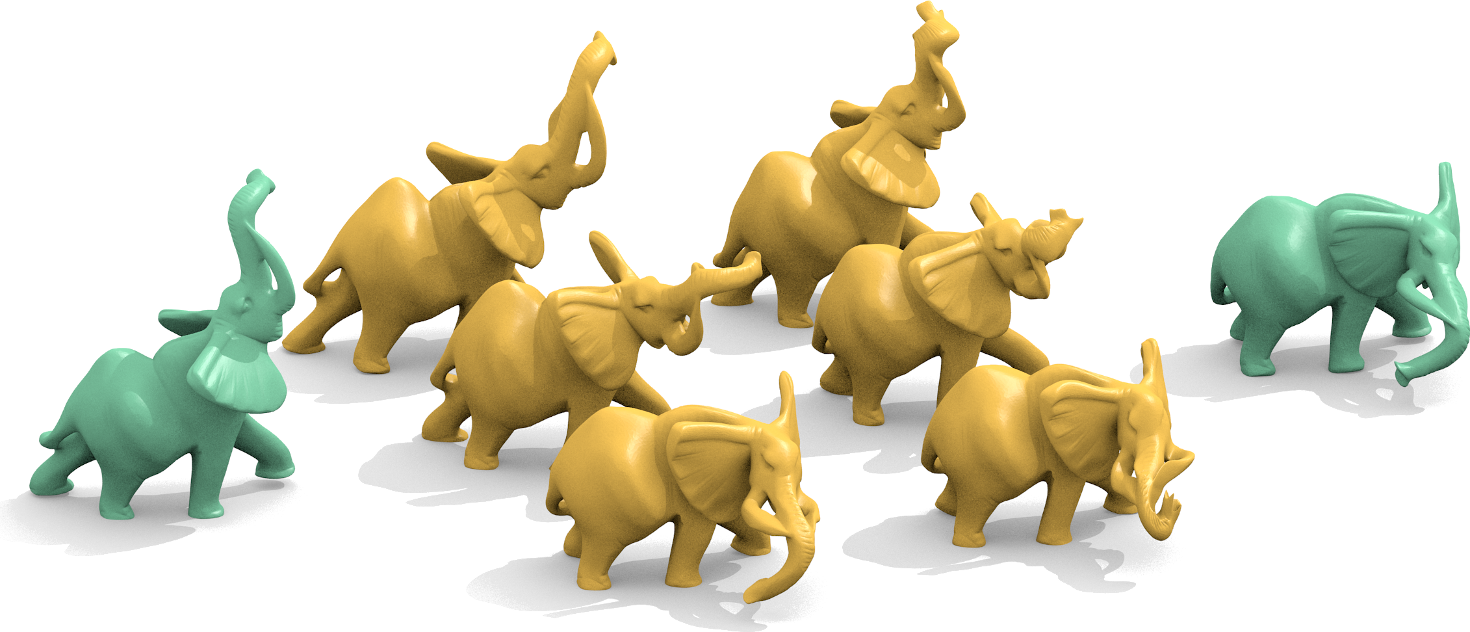}
        \put(29,38){Ours}
        \put(54,38){\cite{eisenberger2019divergence}}
    \end{overpic}
    \caption{A comparison of our method and divergence-free interpolation \cite{eisenberger2019divergence} on a pair of synthetic shapes (green). Both methods preserve the volume but for this large scale deformation, the stationary vector field in \cite{eisenberger2019divergence} is too restrictive which leads to a distorted geometry for $t\geq 0.5$}
    \label{fig:elephants}
\end{figure*}

\begin{figure*}[!b]
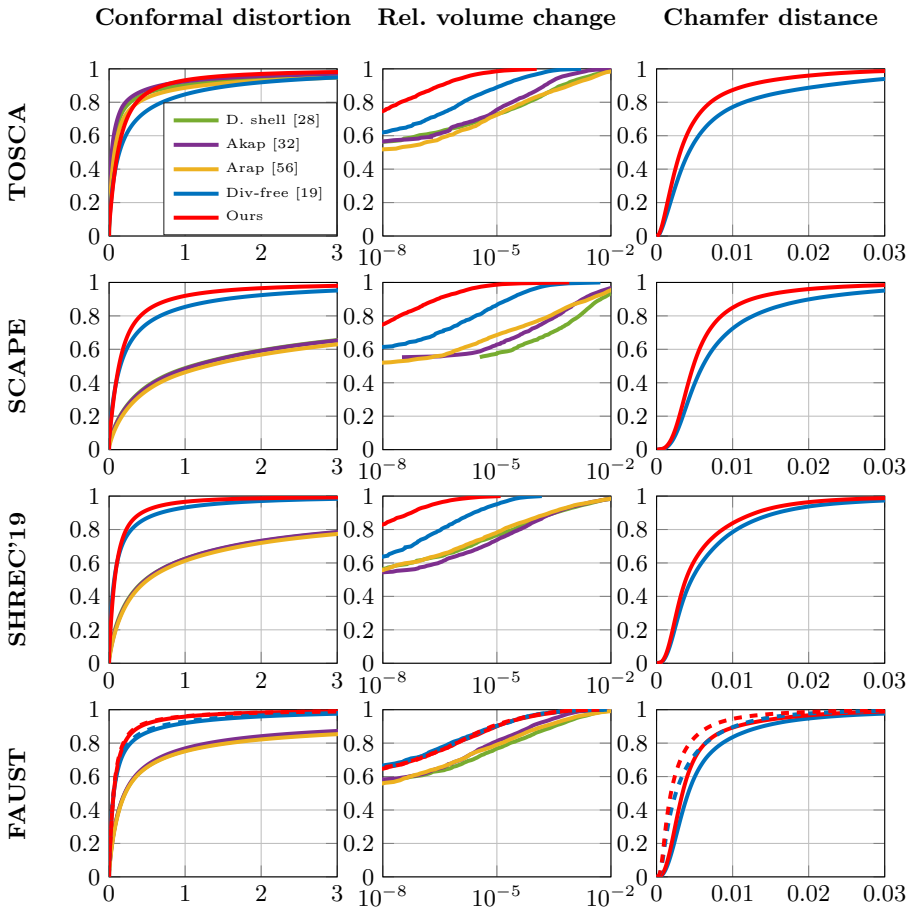

    \centering
    \begin{tikzpicture}
    \input{Figures/curves/TOSCA_0_chamfer_dist.tikz}
    \input{Figures/curves/TOSCA_0_volume_change.tikz}
    \input{Figures/curves/TOSCA_0_conf_dist.tikz}
    
    \input{Figures/curves/SCAPE_chamfer_dist.tikz}
    \input{Figures/curves/SCAPE_volume_change.tikz}
    \input{Figures/curves/SCAPE_conf_dist.tikz}
    
    \input{Figures/curves/SHRECISO_chamfer_dist.tikz}
    \input{Figures/curves/SHRECISO_volume_change.tikz}
    \input{Figures/curves/SHRECISO_conf_dist.tikz}
    
    \input{Figures/curves/FAUST_0_chamfer_dist.tikz}
    \input{Figures/curves/FAUST_0_volume_change.tikz}
    \input{Figures/curves/FAUST_0_conf_dist.tikz}
    \end{tikzpicture}
    \caption{Quantitative results and comparisons with other methods on four benchmarks. For FAUST, dotted lines correspond to the results on the high resolution scans. Those were only computed for our method and \cite{eisenberger2019divergence} because for the other methods the resolution of around 200k vertices is prohibitively high.}
    \label{fig:curves}
\end{figure*}

We verify the generality of our method on four different datasets with increasing complexity. The first two are the synthetic datasets TOSCA \cite{bronstein2008numerical} and SCAPE \cite{anguelov2005scape} where we use the ground truth correspondences for the former and correspondences from BCICP \cite{ren2018orientation} for the latter. The last two datasets SHREC'19 Isometry \cite{dyke2019shrec} and FAUST \cite{Bogo:CVPR:2014} contain reals scans of a puppet and different humans respectively, see Figure~\ref{fig:scans}. For those, we use correspondences from Smooth Shells~\cite{eisenberger2019smooth} and FMNet~\cite{litany2017deep}. Our experiments show that our formulation is applicable to a wide range of inputs with varying levels of noise. Figure~\ref{fig:curves} summarizes our quantitative evaluations on all datasets with comparisons to four other popular interpolation methods. The other methods are Geometric Modeling in Shape Space \cite{kilian2007geometric}, Time-Discrete Geodesics in the Space of Shells \cite{heeren2012time}, Divergence‐Free Shape Correspondence by Deformation \cite{eisenberger2019divergence} and As-Rigid-As-Possible Surface Modeling \cite{sorkine2007rigid}. Although the latter does not describe an interpolation algorithm explicitly, it is trivial to employ its shape deformation procedure in an interpolation pipeline by using Eq.~\eqref{eq:nsequenceenergy}. On the surface, our method is similar to \cite{eisenberger2019divergence} in the sense that both approaches compute divergence-free fields in a low rank basis. The decisive difference is that our method is based on a physically plausible formulation which, among other things, allows for time-dependent vector fields $v(t,x)$. This makes our method more expressive, see Figure~\ref{fig:elephants} for an example.

\paragraph{Error metrics}
In order to quantify the precision of a shape interpolation, we compute three different metrics for each pair of input shapes and plot the resulting cumulative curves in Figure~\ref{fig:curves}. In particular, we measure the conformal distortion \cite[Eq. (6)]{zhang2001efficient} and volume change \cite[Eq. (3)]{hormann2000mips} of intermediate shapes and the Chamfer distance to the target shapes in $\%$ of the diameter for our method and the second alignment based method \cite{eisenberger2019divergence}. If we are strict, the notion of volume change is only meaningful for watertight meshes, which typically does not hold for real scans. 
Our argument regarding this is that in theory, a flow induced by a divergence-free deformation field is exactly volume preserving in terms of the underlying watertight real-world manifold $\mX$. Remarkably, in this way we can even make sense of the notion of volume for a point cloud, assuming that it was sampled from a closed, continuous surface.

\paragraph{Implementation details} 
The low rank vector field representation of divergence-free fields in our Scheme \ref{eq:scheme} is entirely decoupled of the input resolutions $n$ and $m$. Moreover, the vector fields are represented in a spatially dense Eulerian basis which means that at any discrete time $t$, the resulting vector field $\mathrm{v}\bigl(x;c^{(t)}\bigr)$ can be computed for arbitrary points $x$ in our domain, see \cite{eisenberger2019divergence} for more details. This allows us to efficiently perform the optimization in Algorithm~\ref{alg:interpolation} on a subsampled version of the input shapes $p$ and $q$ with a fixed resolution of $2k$ points. Afterwards, the computed vector field can be applied to the full resolution in a single forward pass without any skinning strategy or the like. For once, this makes our approach significantly faster but it also allows for an interpolation of very high resolution objects like those from FAUST ($\sim$200k vertices). Many other classical interpolation methods use some multiscale scheme to allow for higher resolutions \cite{heeren2012time,kilian2007geometric}, but there are still upper limits for them as to what is feasible in terms of computation cost. Our interpolation Algorithm \ref{alg:interpolation} is directly applicable to point clouds, therefore we simply subsample both input shapes using Euclidean farthest point sampling. However, other subsampling strategies like remeshing are also possible if one wants to work directly with meshes. Finally, we use the same set of parameters for all experiments, see our implementation for details.

\begin{figure*}[!b]
    \centering
    \begin{overpic}
        [width=\linewidth]{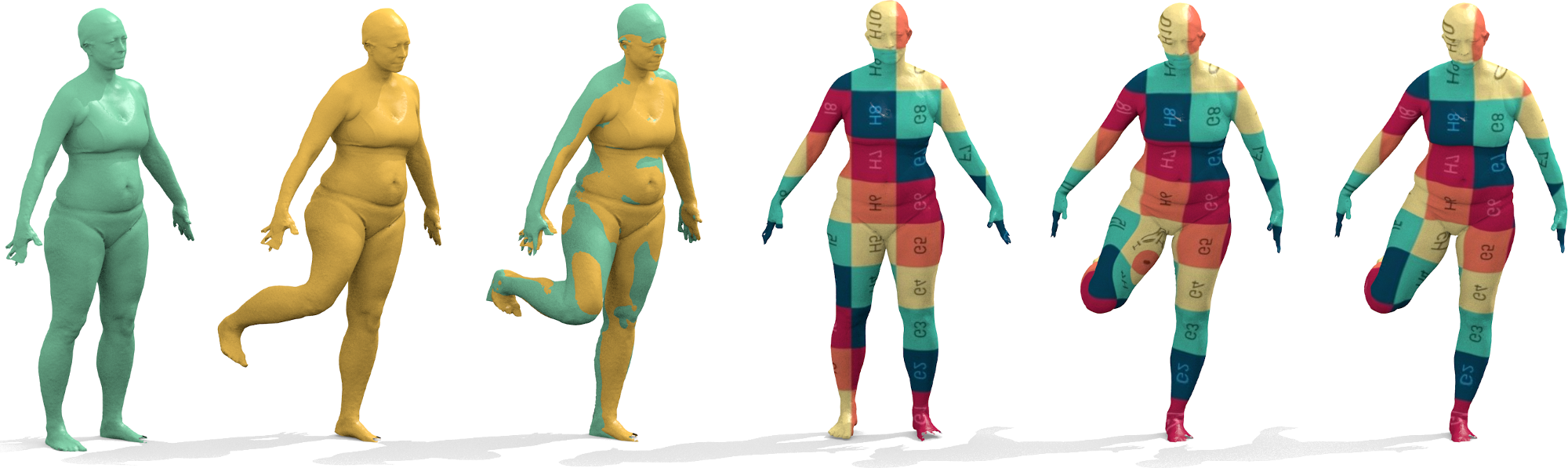}
        \put(73.5,32){\cite{litany2017deep}}
        \put(90.5,32){Ours}
    \end{overpic}
    \caption{We show how our method can be used to refine an imperfect shape correspondence. Using the input matching from Deep Functional Maps \cite{litany2017deep}, we compute an interpolation (left half) and use it to recover the improved correspondences using the final alignment at $t=1$ (right half). We display the matching with a texture map from the first input shape (3rd human from right) to the final pose with both methods.}
    \label{fig:matching_refinement}
\end{figure*}

\begin{figure*}[!b]
    \centering
    \begin{overpic}
        [width=\linewidth]{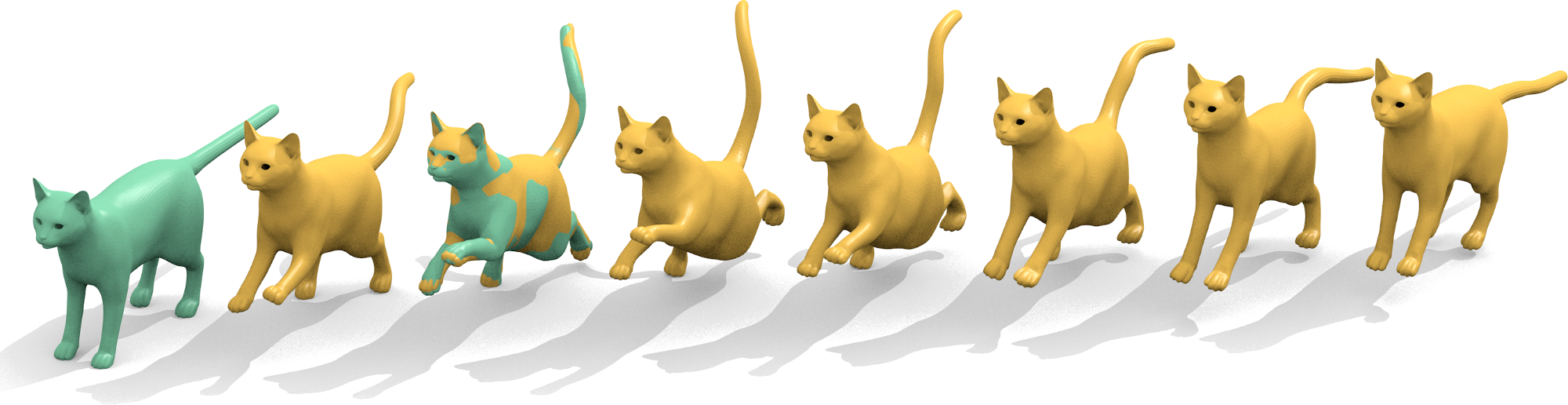}
        \put(5,23){$t=0$\small}
        \put(17,23){$t=0.5$\small}
        \put(29,23){$t=1$\small}
        \put(41,23){$t=2$\small}
        \put(53,23){$t=3$\small}
        \put(65,23){$t=4$\small}
        \put(77,23){$t=5$\small}
        \put(89,23){$t=20$\small}
    \end{overpic}
    \caption{An example of how our approach can be used to extrapolate the motion prescribed by the two input frames $t=0$ and $t=1$. The sequences obtained with our method are physically plausible and remain stable over a long period of time. The cat keeps raising its paw until at $t=2$, driven by the regularizer \eqref{eq:anisotropicarapenergy}, the motion reverses.}
    \label{fig:extrapolate}
\end{figure*}
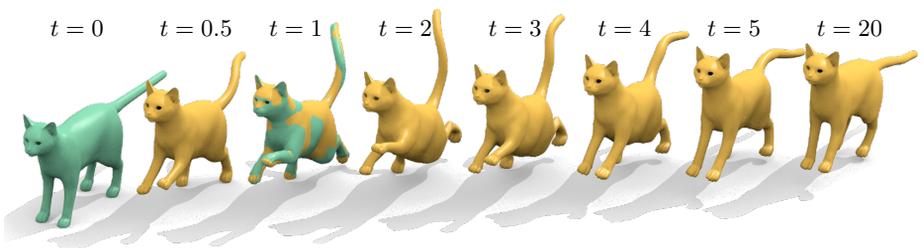

\paragraph{Additional evaluations}

\begin{figure*}
    \centering
    \includegraphics[width=0.95\linewidth]{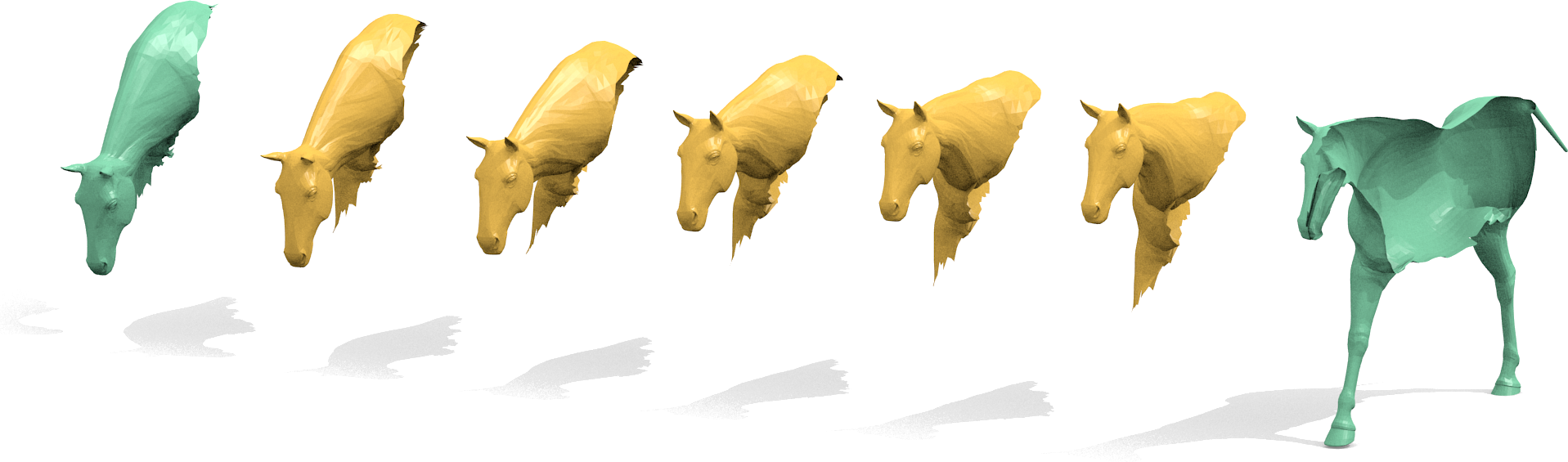}
    \caption{An example interpolation of a pair of partial shapes from the synthetic TOSCA cuts \cite{rodola2017partial} dataset with our method.}
    \label{fig:partial}
\end{figure*}

As a proof of concept, we show that our physically plausible formulation allows for a broad range of applications beyond shape interpolation. For once, we can use our alignment at $t=1$ to refine a shape matching which we show for a real scan of FAUST in Figure~\ref{fig:matching_refinement}. Furthermore, we can compute plausible shape extrapolations by simply simulating the forward integration for a longer period of time than $t=1$. Remarkably, this can be done without any additional optimization, we simply compute an interpolation between $p$ at $t=0$ and $q$ at $t=1$ and then integrate our Scheme~\ref{eq:scheme} until $t>1$, see Figure~\ref{fig:extrapolate}. Finally, we show that our method allows for input objects where only parts of the geometry are available, see Figure~\ref{fig:partial}. This is only feasible for an alignment based method, because the classical formulation as a BVP requires that every vertex has a corresponding point on the other surface. Partial shape interpolation is an important preliminary result for many real world applications like scanning of dynamically moving 3D objects.

\section{Conclusion}

We presented a general and flexible approach to shape interpolation that is systematically derived from a formulation of Hamiltonian dynamics. For this, we employ recent advances in dynamic thin shell simulation to get a robust deformation model and solve its inverse problem by optimizing over the initial motion and anisotropic surface properties. We demonstrated that, in comparison to prior work, our approach is able to compute high quality, physically plausible interpolations of noisy real world inputs. In future work, we will apply our setup to a broader range of applications like 3D scanning of actions or mesh compression.

\clearpage

\bibliographystyle{splncs04}
\bibliography{refs}

\newpage
\appendix

\section{Proof of Theorem~\ref{thm:errororder}}\label{appendix:proof1}
We now provide a proof of Theorem~\ref{thm:errororder} which states that the quality of the approximation of $v^{(t+2)}$ is improved by one error order when the extrapolation step \eqref{eqsub:extrapolate} is added after the velocity update step.
\begin{proof}
    If we remove the extrapolation step \eqref{eqsub:extrapolate} from Algorithm~\ref{alg:interpolation}, Taylor's Theorem implies that $v^{(t+1)}$ yields an estimator of error order $\mathcal{O}(\tau)$ for the velocity in the next timestep $v^{(t+2)}$:
    \begin{equation}
        v^{(t+2)}=v^{(t+1)}+\mathcal{O}(\tau).
    \end{equation}
    The standard backward distance approximation provides an estimation of $\dot{v}^{(t+1)}$:
    \begin{equation}
        \dot{v}^{(t+1)}=\frac{v^{(t+1)}-v^{(t)}}{\tau}+\mathcal{O}(\tau).
    \end{equation}
    Combining this with a Taylor expansion of $v^{(t+2)}$ then yields the statement from the Theorem:
    \begin{flalign}
        v^{(t+2)}=&v^{(t+1)}+\tau\dot{v}^{(t+1)}+\mathcal{O}(\tau^2)=v^{(t+1)}+\tau\frac{v^{(t+1)}-v^{(t)}}{\tau}+\mathcal{O}(\tau^2)=\\&2v^{(t+1)}-v^{(t)}+\mathcal{O}(\tau^2)=\bar{v}^{(t+1)}+\mathcal{O}(\tau^2).
    \end{flalign}
\end{proof}

\section{Runtime analysis}
\begin{figure}[H]
    \centering
%
%
\definecolor{mycolor1}{rgb}{0.00000,0.44700,0.74100}%
\definecolor{mycolor2}{rgb}{0.46600,0.67400,0.18800}%
\definecolor{mycolor3}{rgb}{0.49400,0.18400,0.55600}%
\definecolor{mycolor4}{rgb}{0.92900,0.69400,0.12500}%
\begin{tikzpicture}

\begin{axis}[%
width=0.761\linewidth,
height=0.3\linewidth,
at={(0\linewidth,0\linewidth)},
scale only axis,
xmin=4344,
xmax=52565,
every x tick label/.append style={font=\color{black}, font=\footnotesize},
x tick label style={/pgf/number format/fixed},
scaled x ticks=false,
xtick={10000,20000,30000,40000,50000},
xlabel style={font=\color{white!15!black}},
xlabel={\# points},
ymin=0,
ymax=1600,
ylabel style={font=\color{white!15!black}},
ylabel={time/s},
axis background/.style={fill=white},
xmajorgrids,
ymajorgrids,
legend style={at={(0.20\linewidth,0.295\linewidth)}, legend cell align=left, align=left, draw=white!15!black, font=\tiny}
]

\addplot [color=red, line width=2.0pt]
  table[row sep=crcr]{%
4344	149.236113298684\\
15768	162.361112516373\\
19248	172.777778003365\\
25290	190.677083213814\\
27894	171.98610920459\\
45659	206.729796799746\\
52565	191.527779009193\\
};
\addlegendentry{Ours}

\addplot [color=mycolor1, line width=2.0pt]
  table[row sep=crcr]{%
4344	26.2500043027103\\
15768	35.7777773402631\\
19248	38.8888902962208\\
25290	42.8993051173165\\
27894	44.3333334289491\\
45659	57.0959592746063\\
52565	61.8333339504898\\
};
\addlegendentry{Div-free \cite{eisenberger2019divergence}}

\addplot [color=mycolor2, line width=2.0pt]
  table[row sep=crcr]{%
4344	463.749997783452\\
15768	792.555555514991\\
19248	989.861110690981\\
25290	1215.15624993481\\
27894	1208.56944588013\\
45659	1208.03030338985\\
52565	1341.00555544719\\
};
\addlegendentry{D. shell \cite{heeren2012time}}

\addplot [color=mycolor3, line width=2.0pt]
  table[row sep=crcr]{%
4344	147.29166473262\\
15768	344.166667666286\\
19248	295.416666660458\\
25290	373.333333816845\\
27894	386.2638884224\\
45659	451.199494132941\\
52565	476.466666664928\\
};
\addlegendentry{Akap \cite{kilian2007geometric}}

\addplot [color=mycolor4, line width=2.0pt]
  table[row sep=crcr]{%
4344	122.500000521541\\
15768	309.361111186445\\
19248	272.361110430211\\
25290	351.215275994036\\
27894	407.3611096479\\
45659	545.946969693019\\
52565	566.18333408609\\
};
\addlegendentry{Arap \cite{sorkine2007rigid}}

\addplot [color=red, dashed, line width=2.0pt]
  table[row sep=crcr]{%
4344	141.944447066635\\
15768	142.503014129237\\
19248	141.698308011329\\
25290	163.600922091638\\
27894	144.783165536466\\
45659	180.608520509527\\
52565	166.326398965043\\
};

\addplot [color=red, dashed, line width=2.0pt]
  table[row sep=crcr]{%
4344	156.527779530734\\
15768	182.219210903509\\
19248	203.857247995401\\
25290	217.753244335989\\
27894	199.189052872715\\
45659	232.851073089965\\
52565	216.729159053344\\
};

\addplot [color=mycolor1, dashed, line width=2.0pt]
  table[row sep=crcr]{%
4344	26.2500043027103\\
15768	34.1051009150337\\
19248	37.4190253410868\\
25290	41.4872781295017\\
27894	43.1666670599928\\
45659	56.0728437790004\\
52565	60.9213079398657\\
};

\addplot [color=mycolor1, dashed, line width=2.0pt]
  table[row sep=crcr]{%
4344	26.2500043027103\\
15768	37.4504537654925\\
19248	40.3587552513548\\
25290	44.3113321051314\\
27894	45.4999997979054\\
45659	58.1190747702122\\
52565	62.7453599611139\\
};

\addplot [color=mycolor2, dashed, line width=2.0pt]
  table[row sep=crcr]{%
4344	436.527776531875\\
15768	679.95996432835\\
19248	873.248357819233\\
25290	1163.27783737275\\
27894	1015.82141921745\\
45659	978.426238371067\\
52565	1087.80264241749\\
};

\addplot [color=mycolor2, dashed, line width=2.0pt]
  table[row sep=crcr]{%
4344	490.972219035029\\
15768	905.151146701632\\
19248	1106.47386356273\\
25290	1267.03466249686\\
27894	1401.31747254281\\
45659	1437.63436840863\\
52565	1594.20846847689\\
};

\addplot [color=mycolor3, dashed, line width=2.0pt]
  table[row sep=crcr]{%
4344	130.277775228024\\
15768	327.394778330552\\
19248	264.334090874379\\
25290	345.603687310046\\
27894	353.472316175623\\
45659	422.831998001001\\
52565	441.603130365936\\
};

\addplot [color=mycolor3, dashed, line width=2.0pt]
  table[row sep=crcr]{%
4344	164.305554237217\\
15768	360.938557002021\\
19248	326.499242446537\\
25290	401.062980323644\\
27894	419.055460669177\\
45659	479.566990264881\\
52565	511.33020296392\\
};

\addplot [color=mycolor4, dashed, line width=2.0pt]
  table[row sep=crcr]{%
4344	101.111110299826\\
15768	291.727666698881\\
19248	248.476260561082\\
25290	320.656898777022\\
27894	348.009351754668\\
45659	511.274135249967\\
52565	529.768836817662\\
};

\addplot [color=mycolor4, dashed, line width=2.0pt]
  table[row sep=crcr]{%
4344	143.888890743256\\
15768	326.994555674009\\
19248	296.245960299339\\
25290	381.773653211051\\
27894	466.712867541132\\
45659	580.61980413607\\
52565	602.597831354519\\
};

\end{axis}
\end{tikzpicture}%
    \caption{A runtime analysis of different interpolation methods on all pairs in TOSCA. We plot the mean computation time (solid line) and two lines corresponding to one standard deviation (dashed).}
    \label{fig:runtime}
\end{figure}
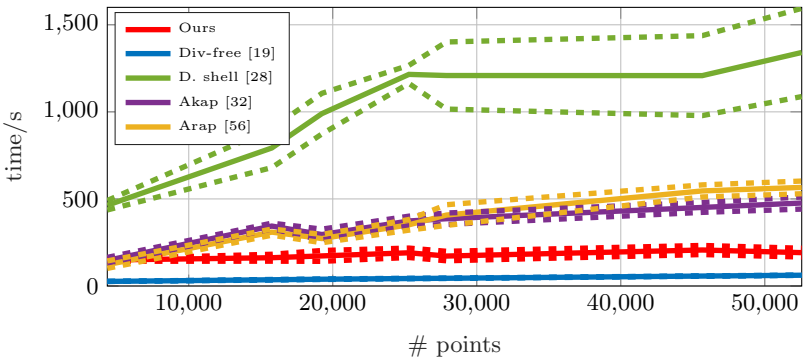

We compare the runtime of our method to other popular shape interpolation methods based on our experiments on TOSCA in Figure~\ref{fig:runtime}. Only divergence-free interpolation \cite{eisenberger2019divergence} is faster than our method. Most importantly, for our approach and \cite{eisenberger2019divergence} the runtime is essentially independent of the resolution because the optimization is done on a fixed resolution of 2k vertices. Only the last forward pass on the whole input shape $p\in\bbR^{n\times 3}$ depends on the resolution $n$ but this step is cheap in comparison to the optimization.

\section{Additional qualitative evaluations}

\begin{figure*}
    \centering
    \includegraphics[width=0.85\linewidth]{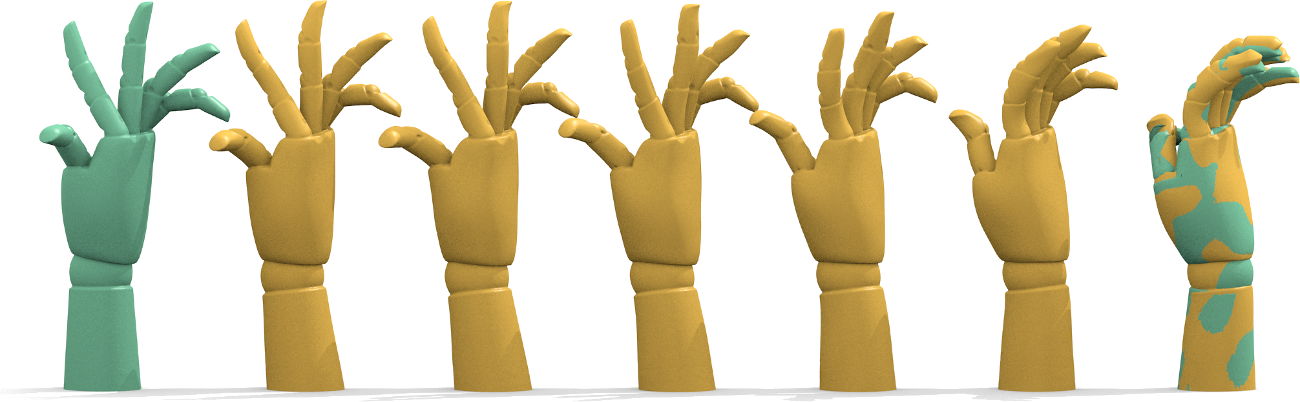}
    \includegraphics[width=0.85\linewidth]{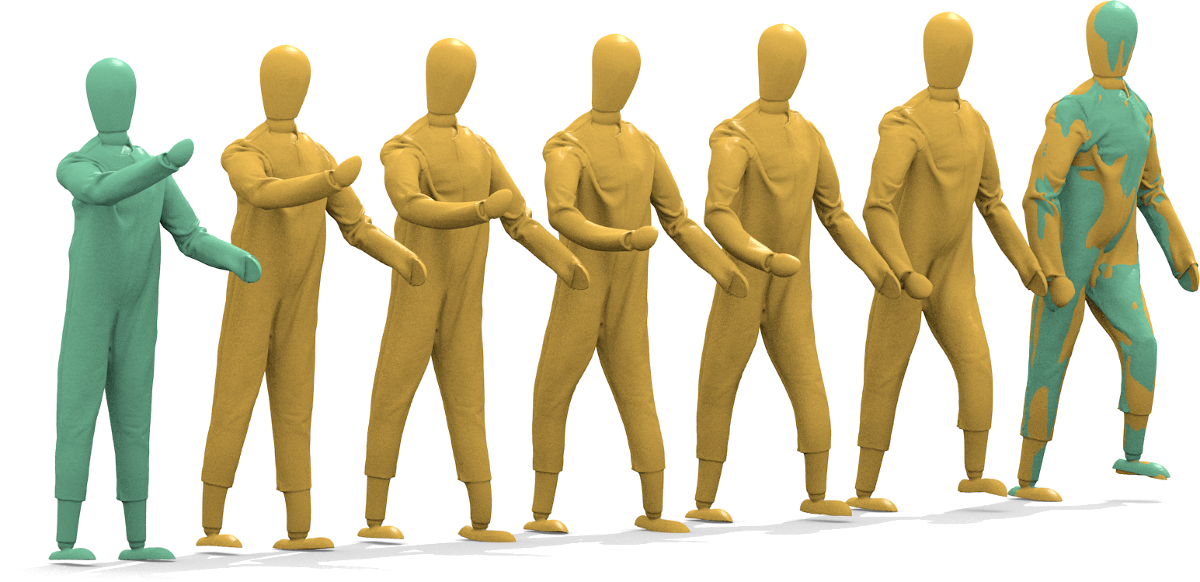}
    \includegraphics[width=0.85\linewidth]{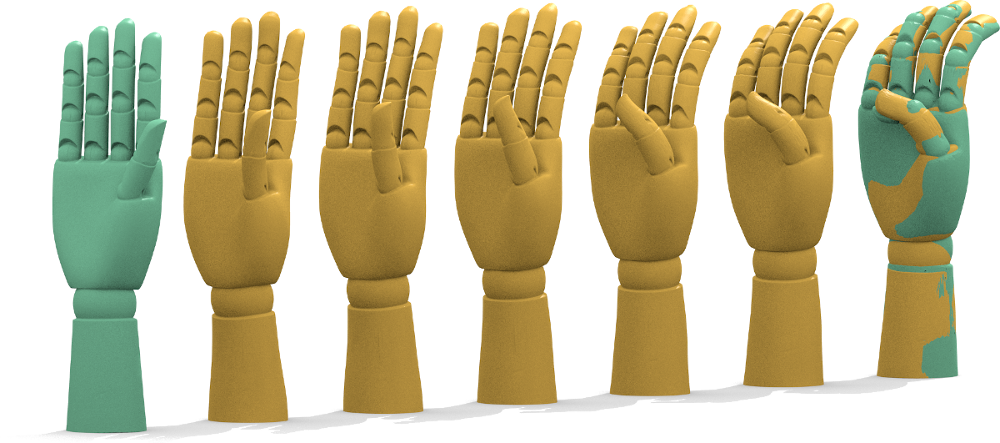}
    \caption{Additional examples of interpolations with our method on SHREC'19.}
    \label{fig:additionalqualshreciso}
\end{figure*}

\begin{figure*}[t!]
    \centering
    \includegraphics[width=0.95\linewidth]{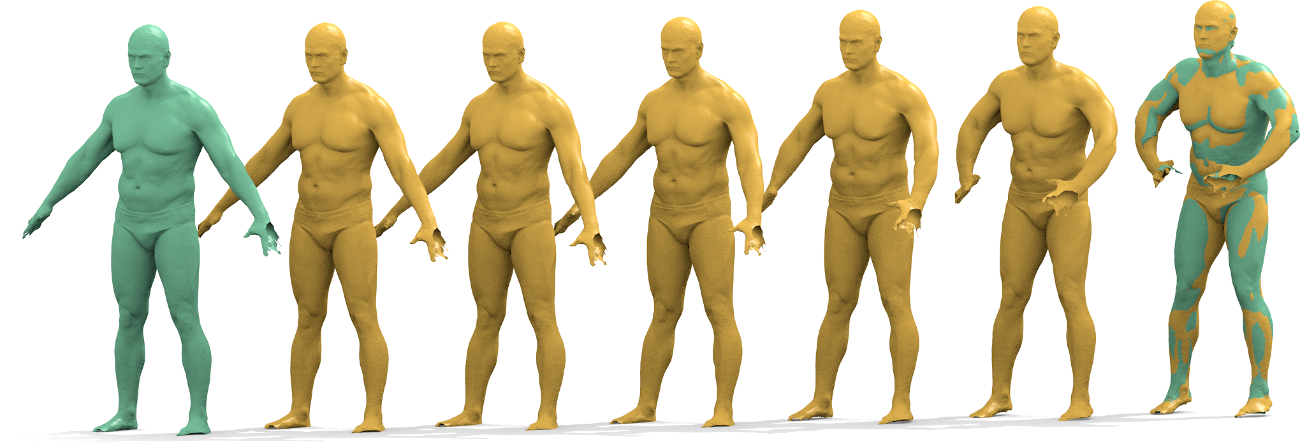}
    \includegraphics[width=0.95\linewidth]{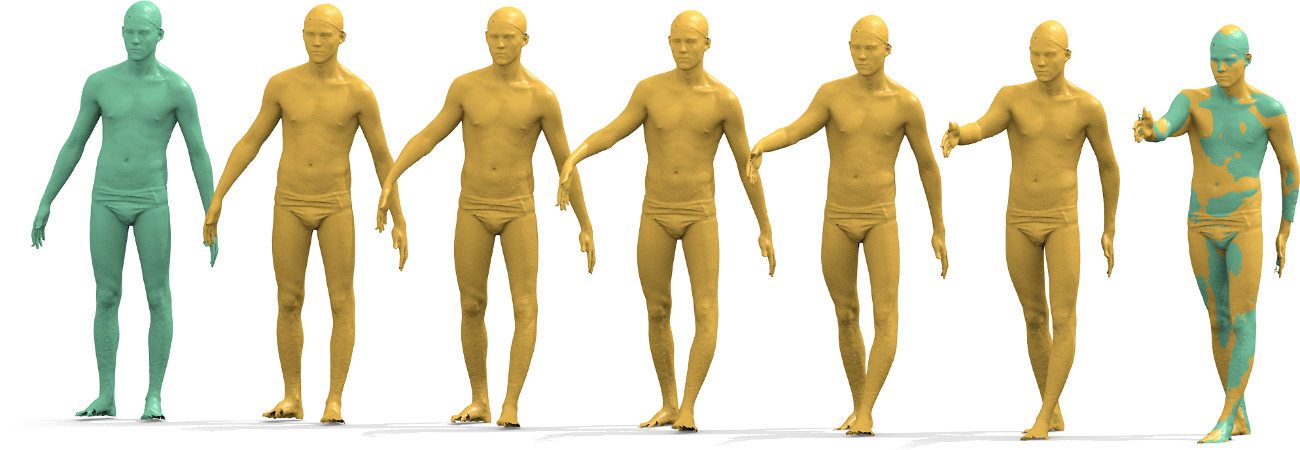}
    \caption{Additional examples of interpolations with our method on FAUST.}
    \label{fig:additionalqualfaust}
\end{figure*}

\begin{figure*}[!b]
    \centering
    \begin{overpic}
        [width=0.8\linewidth]{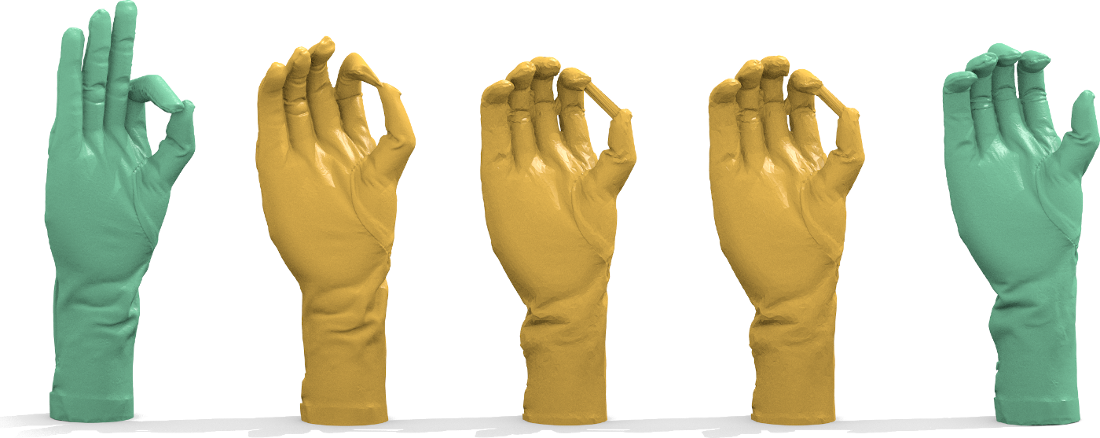}
        \put(26,38){Ours}
        \put(47,38){\cite{kilian2007geometric}}
        \put(68,38){\cite{heeren2012time}}
    \end{overpic}
    \caption{An example of a scanned hand from SHREC'19 Isometry \cite{dyke2019shrec} where we show the second to last frame from our method, \cite{heeren2012time} and \cite{kilian2007geometric} respectively. In SHREC'19 there are various pairs with topological changes. In the case presented here, the meshing connects between the index finger and the thumb. This makes our method fail because the input matching from \cite{eisenberger2019smooth} is not able to separate the two fingers entirely.}
    \label{fig:topology}
\end{figure*}

To give a more complete picture, we show additional examples of interpolations with our method on the two datasets with reals scans, SHREC'19 Isometry and FAUST in Figure~\ref{fig:additionalqualshreciso} and Figure~\ref{fig:additionalqualfaust}. Finally, we show a failure case under topological changes on a scanned hand from SHREC'19 in Figure~\ref{fig:topology}. Although our method is not able to separate the touching parts for these cases, our method still produces a more plausible result than other classical interpolation methods. In our case, the fingers appear to be glued together whereas \cite{heeren2012time} produces undesirable artifacts.

\end{document}